\documentclass[conference]{IEEEtran}
\usepackage{algorithmic}
\usepackage{amsmath,amssymb,amsfonts}
\usepackage{cite}
\usepackage{graphicx}
\usepackage{hyperref}
\usepackage{makecell}
\usepackage{multirow}
\usepackage{orcidlink}
\usepackage{textcomp}
\usepackage{url}
\usepackage{tikz}
\usepackage{xcolor}
\usepackage{xspace}
\usepackage[super]{nth}
\def\BibTeX{{\rm B\kern-.05em{\sc i\kern-.025em b}\kern-.08em
    T\kern-.1667em\lower.7ex\hbox{E}\kern-.125emX}}

\graphicspath{{imgs}}

\newcommand{\AIname}{ZeusAI\xspace}
\begin{document}

\title{Learning to Play 7 Wonders Duel\\ Without Human Supervision
}

\author{
    \IEEEauthorblockN{Giovanni Paolini}
    \IEEEauthorblockA{%
        \textit{University of Bologna}\\
        Bologna, Italy\\
        g.paolini@unibo.it\\
        \orcidlink{0000-0002-3964-9101}
        0000-0002-3964-9101
    } \and
    \IEEEauthorblockN{Lorenzo Moreschini}
    \IEEEauthorblockA{%
        \textit{University of Brescia}\\
        Brescia, Italy\\
        lorenzo.moreschini@unibs.it\\
        \orcidlink{0000-0002-4203-3359}
        0000-0002-4203-3359
    } \and
        \IEEEauthorblockN{Francesco Veneziano}
        \IEEEauthorblockA{%
        \textit{University of Genoa}\\
        Genoa, Italy\\
        veneziano@dima.unige.it\\
        \orcidlink{0000-0002-2225-7769} 0000-0002-2225-7769
    } \and
        \IEEEauthorblockN{Alessandro Iraci}
        \IEEEauthorblockA{%
        \textit{University of Pisa}\\
        Pisa, Italy\\
        alessandro.iraci@unipi.it\\
        \orcidlink{0000-0002-3158-3929} 0000-0002-3158-3929
    }
}

\IEEEoverridecommandlockouts

\IEEEpubid{\makebox[\columnwidth]{
\begin{minipage}{\columnwidth}
    \copyright2024 IEEE. Personal use of this material is permitted. Permission from IEEE must be obtained for all other uses, in any current or future media, including reprinting/republishing this material for advertising or promotional purposes, creating new collective works, for resale or redistribution to servers or lists, or reuse of any copyrighted component of this work in other works.
\end{minipage}
\hfill}
\hspace{\columnsep}\makebox[\columnwidth]{ }}

\maketitle

\IEEEpubidadjcol

\pagestyle{plain}

\begin{abstract}

This paper introduces ZeusAI, an artificial intelligence system developed to play the board game 7 Wonders Duel.
Inspired by the AlphaZero reinforcement learning algorithm, ZeusAI relies on a combination of Monte Carlo Tree Search and a Transformer Neural Network to learn the game without human supervision.
ZeusAI competes at the level of top human players, develops both known and novel strategies, and allows us to test rule variants to improve the game's balance.
This work demonstrates how AI can help in understanding and enhancing board games.
\end{abstract}


\section{Introduction}
\label{sec:introduction}

7 Wonders Duel (7WD) is an award-winning board game published in 2015, as a 2-player version of the world-famous multi-player game 7 Wonders.
As of May 2024, 7WD holds the 19th place in the BoardGameGeek ranking,\footnote{\url{https://boardgamegeek.com/browse/boardgame}}
and is the 8th most played game on Board Game Arena (BGA)\footnote{\url{https://boardgamearena.com}} with 8.7 million games played since its introduction in 2020.
Its presence on BGA has fostered competitive play and the development of deep strategies, also thanks to frequent tournaments as well as seasonal and all-time rankings.

This paper introduces \AIname, an AI system capable of matching, and potentially outperforming, top human players at 7WD.
To the best of our knowledge, this is the first publicly declared AI that plays 7WD at an expert level.
We train \AIname using Deep Reinforcement Learning, with no supervision from games played between humans or from human-developed strategies.
Our approach is similar to AlphaZero \cite{silver2018general}, the algorithm developed by DeepMind to achieve super-human performance in games such as Go, Chess, and Shogi.

\AIname differs from AlphaZero in two key aspects.
First, it has a mechanism to bound the number of random continuations considered when making a move (this is not needed in AlphaZero, as games like Go and Chess have no randomness).
Second, it employs a Transformer \cite{vaswani2017attention} in place of a Convolutional Neural Network.
Transformers have revolutionized natural language processing and, more recently, the whole field of AI \cite{chatgpt}; unlike Convolutional and Recurrent Networks, Transformers do not have a built-in assumption of spatial or temporal proximity in the input, making them better-suited to handle the complex state of a 7WD game.

We are currently in the process of testing \AIname against some of the best players in the world.
To this end, we have been holding best-of-5 live matches\footnote{
These matches are streamed live on YouTube at \url{https://www.youtube.com/playlist?list=PLA4iH7nzdJ7sNWxqAn2JxB_zQ0nEdoJeg}
} as well as an ongoing series of offline games.
\AIname has won 26 out of the 38 games (68.4\%) played so far against top human players.

\AIname rediscovered well-known strategies to play 7WD, but it also frequently makes moves that are counter-intuitive to expert players, suggesting that 7WD is far from solved.
We use \AIname to derive numerous statistics of interest, such as a ranking of the wonders and progress tokens.
We also confirm the known belief that the \nth{1} player holds a significant advantage, estimating a 66.8\% win rate.
We propose and evaluate several variants that reduce this imbalance to as little as 51.6\%, while keeping changes to a minimum; we hope that these variants will gain traction in the 7WD community.

\section{The Game of 7 Wonders Duel}

7WD is a game with no hidden information, and so it admits a theoretically optimal strategy (like Go and Chess) with no dynamics such as bluffing.
However, unlike Go and Chess, 7WD has elements of randomness.

In 7WD, 2 players compete to develop their cities and win in one of 3 ways: civilian (by obtaining the most points), scientific (by acquiring 6 distinct scientific symbols), or military (by advancing the conflict pawn to the opponent's capital on the military track).
These different victory conditions introduce considerable strategic depth, as players need to carefully balance all of them.

The game components include 10 progress tokens, 12 wonders, and 73 building cards, all granting permanent bonuses or one-time effects.
Notably, 5 of the wonders grant the owner an immediate extra turn when constructed.

During game setup, 5 progress tokens are randomly revealed, and then the \emph{wonder selection phase} begins, consisting of two \emph{drafts}.
In the first draft, 4 wonders are randomly revealed and selected following the \nth{1}-\nth{2}-\nth{2}-\nth{1} player sequence.
In the second draft, 4 more wonders are revealed and selected according to the \nth{2}-\nth{1}-\nth{1}-\nth{2} sequence.
This way, each player obtains 4 wonders to build throughout the~game.

\begin{figure}[tb]
    \centerline{\includegraphics[height=2.5cm]{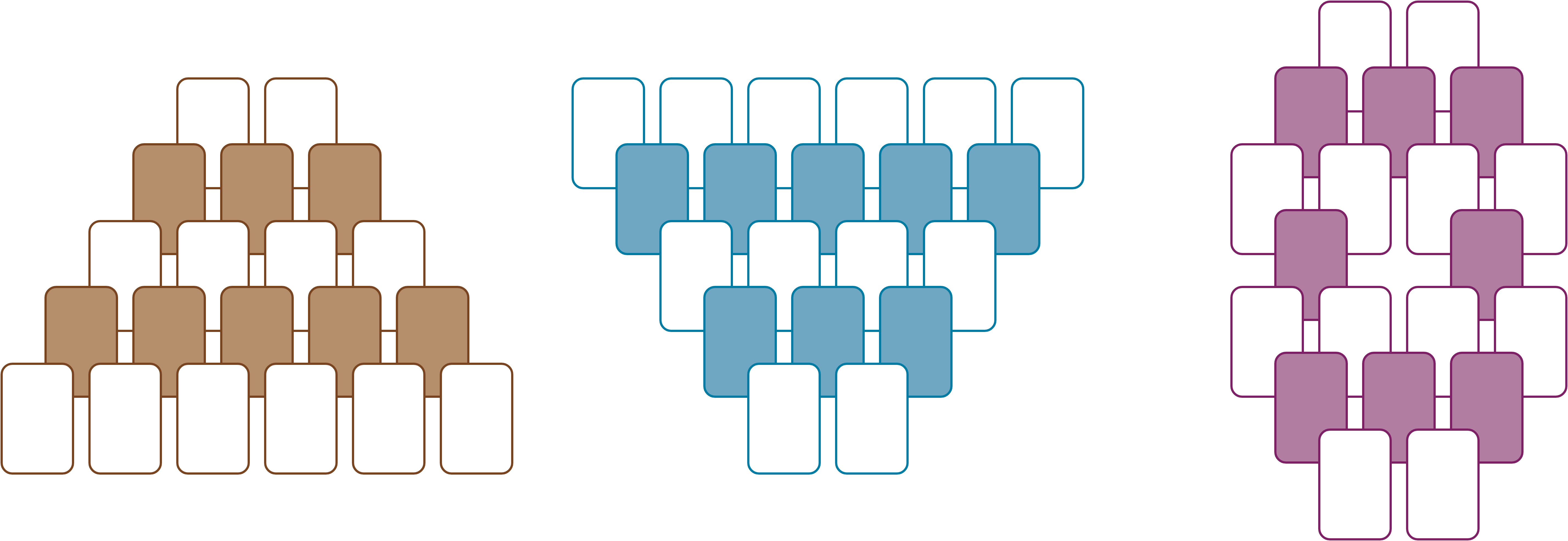}}
    \caption{Structure of face-up cards (white) and face-down cards (colored) in ages I, II, and III (from left to right).}
    \label{fig:age_setup}
\end{figure}

\begin{figure}[tb]        
    \centerline{\includegraphics[width=.48\textwidth]{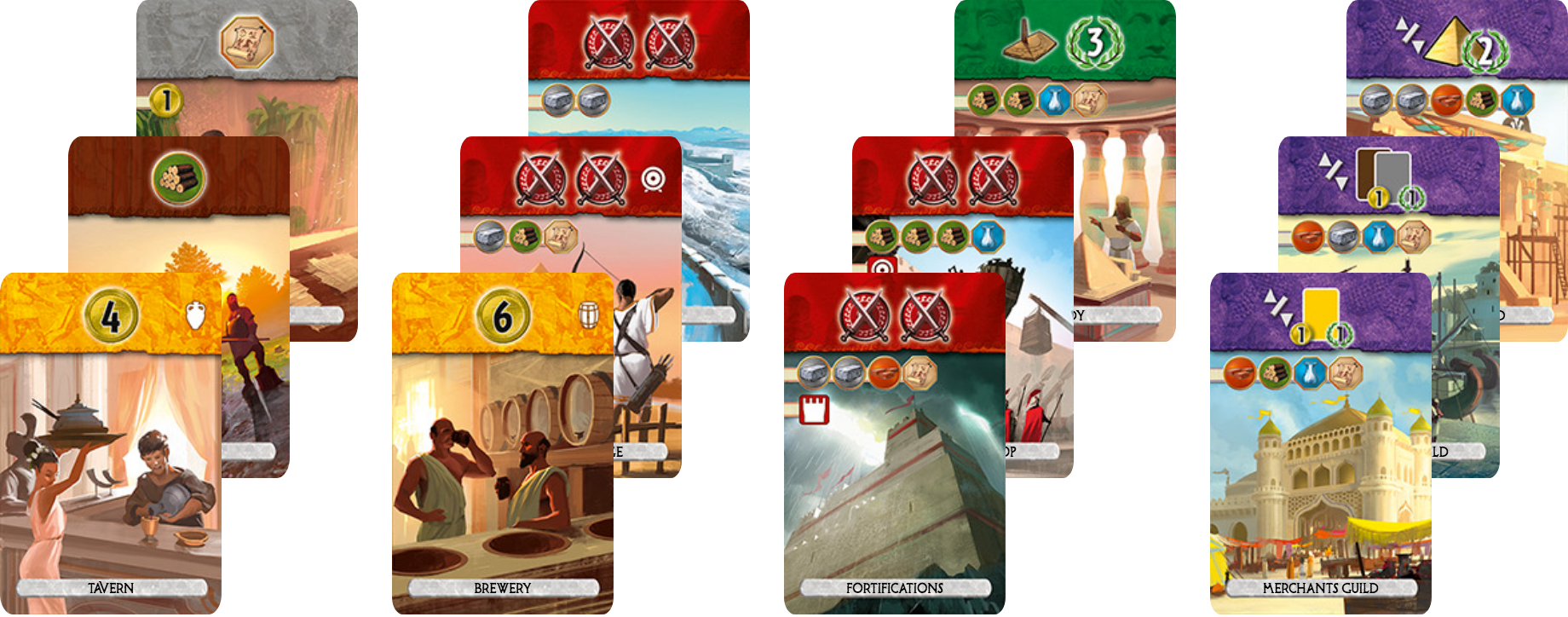}}
     \caption{The 3 most frequently built cards by \AIname in ages I, II, III, and among the guilds.
     In every group, the most frequently built card is shown in front.}
     \label{fig:top_cards}
 \end{figure}

Gameplay progresses through 3 ages, each involving a card selection mechanism from a unique structure of face-up and face-down cards (see Fig.~\ref{fig:age_setup}).
A card can only be selected if it is not covered by any other, with face-down cards being revealed once all cards on top of them have been removed.
Players collect various types of cards to build their cities: brown and grey cards for resources, yellow for commercial advantages, red for military power, green for scientific symbols, blue and purple for victory points (purple cards are called \emph{guilds}).
Some of the cards are shown in Fig.~\ref{fig:top_cards}.

On each turn, a player must choose a card and either: pay its cost to add it to their city and gain its benefits; use it to activate one of their wonders; or discard it to gain coins (which in turn can be used to purchase resources). %
When a player collects two identical scientific symbols, they can pick one of the available progress tokens.

Only 7 wonders can be built in total.
This rule adds pressure to the players, as it is typically convenient to construct all 4 available wonders before the opponent does.
Some of the most critical decisions in a game revolve around finding the correct timing to build the wonders.

\section{\AIname}

\AIname is an adaptation of the celebrated AlphaZero reinforcement learning algorithm \cite{silver2018general}, combining Monte Carlo Tree Search (MCTS) with a Deep Neural Network that models state values and policy.
In this section, we give a technical overview of \AIname.
Some details are omitted for brevity, and we defer a complete description to a future paper.

\subsection{The Model}

In 7 Wonders Duel, unlike Go and Chess, the state does not have a chessboard-like geometry, making Convolutional Neural Networks not well-suited.
Instead, we use Transformer Encoder models \cite{vaswani2017attention} of different sizes.
Our final model has 12 layers and 12 attention heads; it maintains 768-dimensional representations, and the feedforward layers have a hidden dimension of 3072.
A state is represented as a set of components (cards, wonders, progress tokens, etc.), each paired with their positional information (e.g., a card can be in a specific location of the age structure, in either player's city, discarded, or used to build a wonder).
All components and positions have a learned 768-dimensional embedding, akin to words and positions in Transformer-based language models.
No explicit information about the rules, and the costs and effects of game components, is encoded; it is all learned through self-play. Additional tokens encode other aspects of the state, such as the number of coins and the status of the military track.

ZeusAI's Transformer outputs a state value between $-1$ and $+1$, representing the odds of the current player winning the game, and a probability distribution over all legal actions (the policy), with more promising actions having higher probabilities.
The total number of trainable parameters is $\sim$92M.

\subsection{Action Selection}

To select an action in a given state, ZeusAI employs the MCTS algorithm.
Specifically, the procedure begins with a tree consisting only of the root node, representing the current state. During each simulation, the tree is traversed starting from the root, until a new state is encountered and added; actions are chosen in a way that balances exploration (prioritizing less explored actions), and exploitation (favoring actions considered more promising by the model or that have previously led to high-value outcomes).
In some cases, applying an action results in a state where a random event, such as revealing one or more cards, must occur. We refer to these states as \emph{afterstates}.
To reduce the branching factor of the search tree, each afterstate is limited to having at most 11 child states; in non-training games, this limit is gradually increased as the number of visits to an afterstate increases.
If a simulation reaches an afterstate already at the limit, it continues randomly with one of the existing child states. The limit of 11 is based on the maximum number of child states possible when a single card must be revealed. %

When generating self-play games for training, we run MCTS for up to 1k simulations; the action is selected randomly, weighted by the number of visits.
In non-training games, we run up to 5k simulations and deterministically select the action with the highest number of visits.

\subsection{Training}

We initially generated 35k games using three simple rule-based policies: the first policy selects actions at random, with a preference for not discarding; the second and third policies attempt to build random green and red cards, which are required to achieve scientific and military victories respectively, and otherwise behave as the first policy.
We trained the value head of our Transformer model on all states from these games, assigning a target value based on the game's outcome ($+1$ for a win, $-1$ for a loss, and $0$ for a draw).

Subsequent self-play games were generated with ZeusAI.
Every $\sim$3k games, the Transformer model was re-trained on the states from the most recent $\sim$100k games. Specifically, the policy head was trained to predict the visit distribution of the legal actions at the MCTS root node.
We generated a total of $\sim$420k self-play games for training, which would take 9 years of real-time training for a human player on BGA. %

\section{Results}

\subsection{Analysis of the Training Games}

\begin{figure}[tbp]
    \centerline{\includegraphics[width=0.48\textwidth]{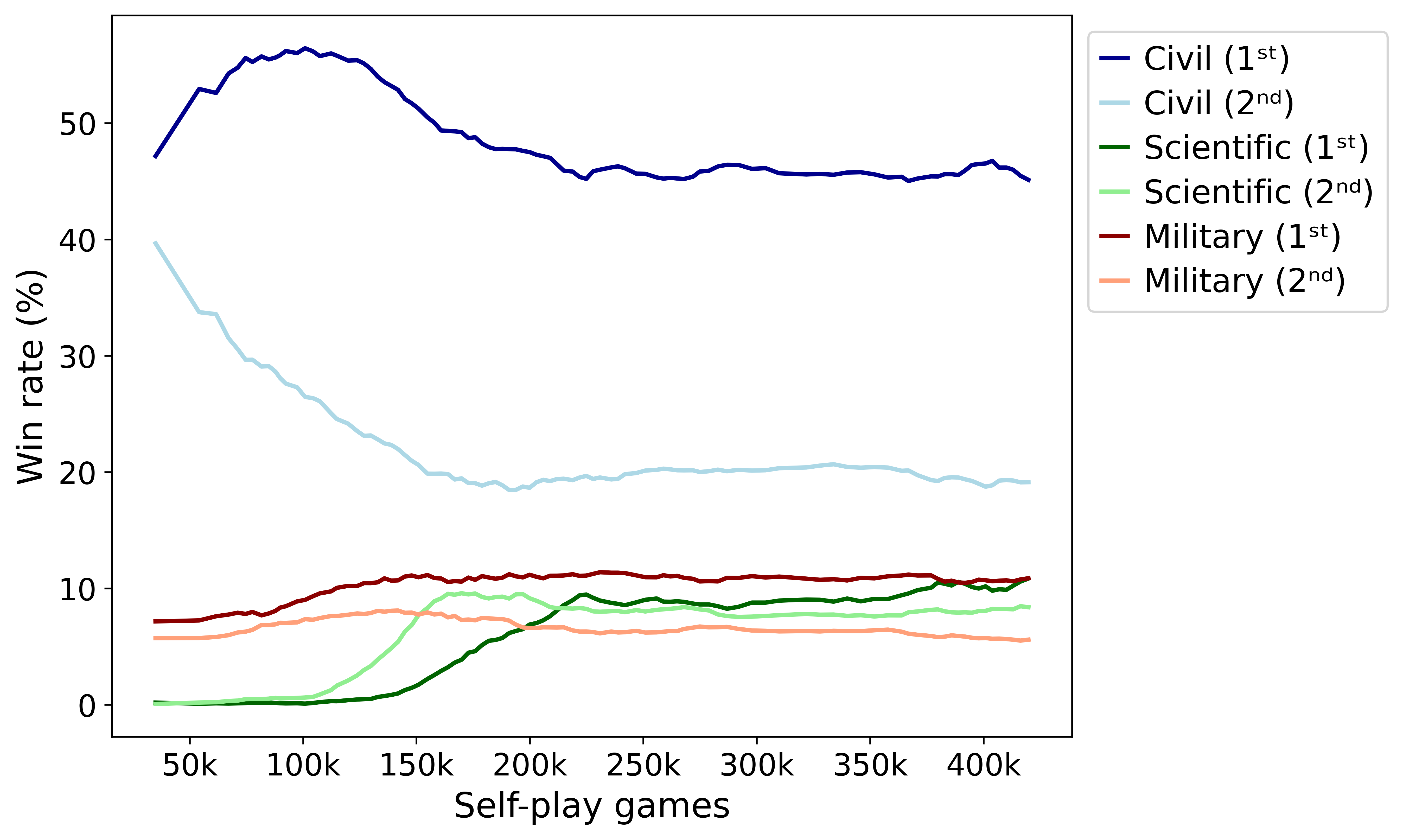}}
    \caption{Distribution of victory types and winners (\nth{1} or \nth{2} player) in self-play games throughout training.}
    \label{fig:victory-type-plot}
\end{figure}

\begin{figure}[tbp]
    \centerline{\includegraphics[width=0.48\textwidth]{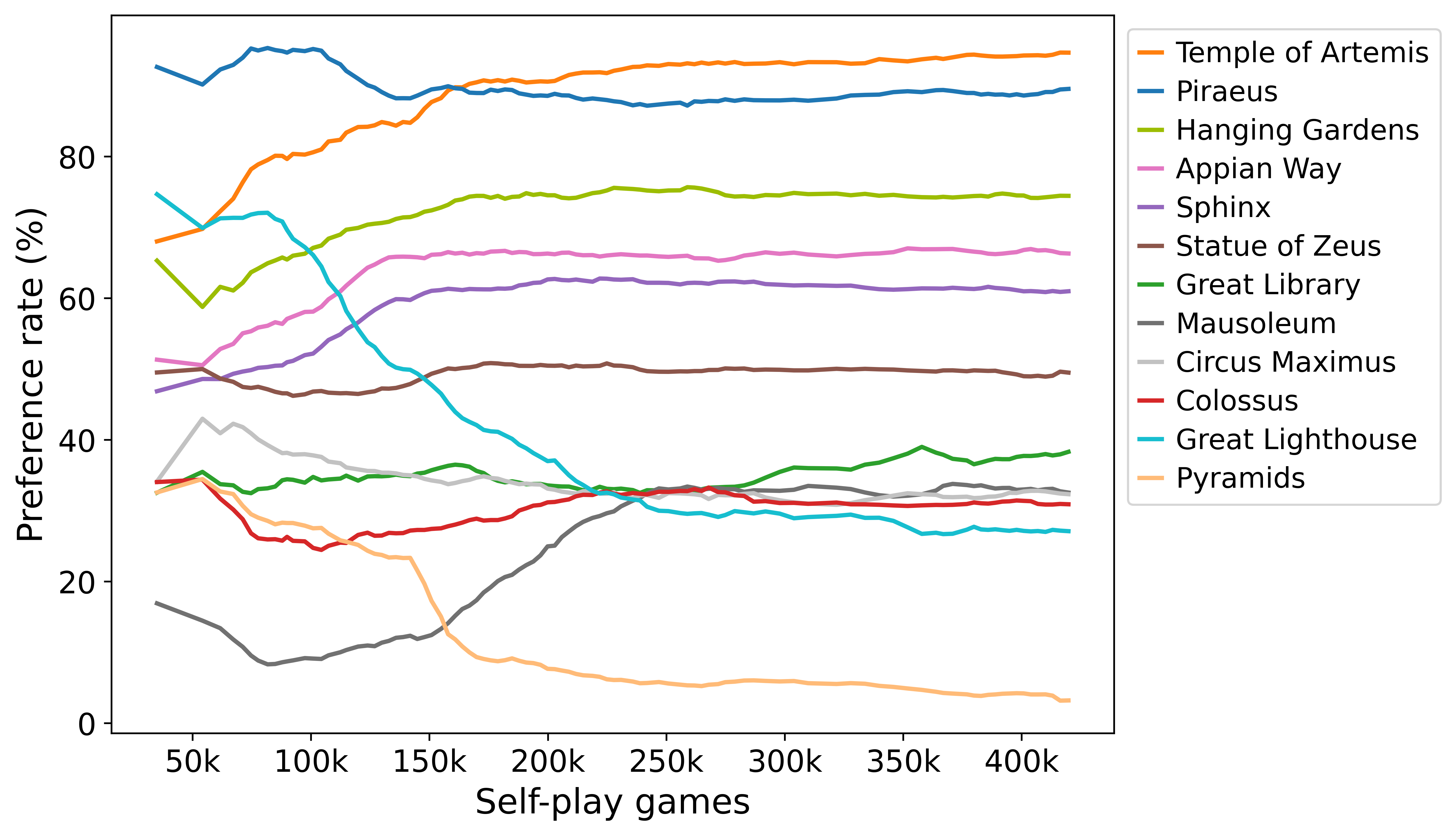}}
    \caption{Wonder preference in self-play games throughout training.
    }
    \label{fig:wonder-preference-plot}
\end{figure}

Fig.~\ref{fig:victory-type-plot} illustrates the evolution of victory types and winners in the self-play games generated by the various iterations of ZeusAI, as a function of the cumulative number of games up to each iteration. Draws only occurred in about 0.1\% of the games and are not shown.

\AIname quickly learned to exploit the \nth{1} player advantage, but it took $\sim$100k games for the scientific victory to be ``discovered'' as a viable strategy, due to the very sparse reward obtained when collecting a diverse set of science symbols.
The discovery of the scientific victory also marked the moment when we stopped being able to consistently beat \AIname and felt confident that it had reached a competent level of play.

Fig.~\ref{fig:wonder-preference-plot} shows the evolution of \AIname's wonder preference throughout training, computed averaging a numerical value based on whether the wonder was selected first (100\%), second or third (50\%), or fourth (0\%) during the draft.
The Pyramids and the Great Lighthouse became less preferred as \AIname learned to exploit wonders offering more strategic benefits.

\subsection{Games Against Human Players}

We turned to the BGA community to further assess \AIname's level of play.
Between January and April 2024, we organized best-of-5 challenges against 3 of the top 7WD players on BGA.
These events were streamed live on YouTube and accompanied by our analysis of the games (the link is provided in Section \ref{sec:introduction}).
Two of the challenges ended with \AIname's victory (3--2 and 3--1) and one with a defeat (2--3).
We also started a separate series of games against one of the three players, and \AIname is currently leading 18--6.
Overall, \AIname has won 26 out of all the 38 games played so far against top players, confirming the impression that it has achieved a very high level of play.

\subsection{Strategic Insights}

\textbf{Victory types:}
After training \AIname, we generated 10k self-play games to extract statistics and insights.
The distribution of victory types is 61.7\% civilian / 21.4\% scientific / 16.9\% military, which is not far from the 58.0\% / 25.6\% / 16.4\% distribution compiled from games played by good players on BGA \cite{bga}.

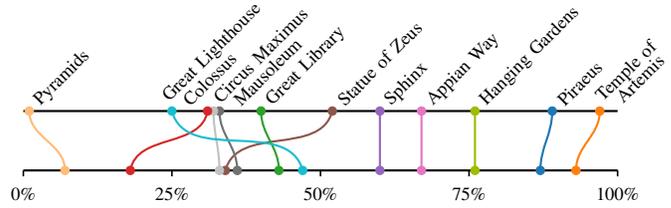
\begin{figure}
    \centering
    \begin{tikzpicture}[scale=0.79]
        \definecolor{TempleOfArtemis}{HTML}{ff7f0e}
        \definecolor{Piraeus}{HTML}{1f77b4}
        \definecolor{HangingGardens}{HTML}{9cbd02}
        \definecolor{AppianWay}{HTML}{e377c2}
        \definecolor{Sphinx}{HTML}{9467bd}
        \definecolor{StatueOfZeus}{HTML}{8c564b}
        \definecolor{GreatLibrary}{HTML}{2ca02c}
        \definecolor{Mausoleum}{HTML}{717171}
        \definecolor{CircusMaximus}{HTML}{c2c2c2}
        \definecolor{Colossus}{HTML}{d62728}
        \definecolor{GreatLighthouse}{HTML}{17becf}
        \definecolor{Pyramids}{HTML}{ffbb78}
    
        \draw[thick] (0,2) -- (10,2);
        \draw[thick] (0,1) -- (10,1);

        \draw[semithick] (0,1) -- (0,0.9);
        \draw[semithick] (2.5,1) -- (2.5,0.9);
        \draw[semithick] (5,1) -- (5,0.9);
        \draw[semithick] (7.5,1) -- (7.5,0.9);
        \draw[semithick] (10,1) -- (10,0.9);

        \scriptsize
        \node at (0,.6) {0\%};
        \node at (2.5,.6) {25\%};
        \node at (5.0,.6) {50\%};
        \node at (7.5,.6) {75\%};
        \node at (10,.6) {100\%};

        \newcommand{\circlesize}{2pt}

        \draw[TempleOfArtemis] (9.5, 2) node[rotate=45, right] {\color{black}{\;\;Temple of}};
        \filldraw[TempleOfArtemis] (9.7, 2) circle (\circlesize) node[rotate=45, right] (T1) {};
        \node[rotate=45] at (10.38, 2.55) {Artemis};
        
        \filldraw[Piraeus] (8.9, 2) circle (\circlesize) node[rotate=45, right] (P1) {\color{black}{\;Piraeus}};
        \filldraw[HangingGardens] (7.6, 2) circle (\circlesize) node[rotate=45, right] (H1) {\color{black}{\;Hanging Gardens}};
        \filldraw[AppianWay] (6.7, 2) circle (\circlesize) node[rotate=45, right] (A1) {\color{black}{\;Appian Way}};
        \filldraw[Sphinx] (6.0, 2) circle (\circlesize) node[rotate=45, right] (S1) {\color{black}{\;Sphinx}};
        \filldraw[StatueOfZeus] (5.2, 2) circle (\circlesize) node[rotate=45, right] (Z1) {\color{black}{\;Statue of Zeus}};
        \filldraw[GreatLibrary] (4.0, 2) circle (\circlesize) node[rotate=45, right] (B1) {\color{black}{\;Great Library}};
        
        \draw[Mausoleum] (3.45, 2) node[rotate=45, right] {\color{black}{\;Mausoleum}};
        \filldraw[Mausoleum] (3.3, 2) circle (\circlesize) node (M1) {};
        
        \draw[CircusMaximus] (3.05, 2) node[rotate=45, right] {\color{black}{\;\;Circus Maximus}};
        \filldraw[CircusMaximus] (3.2, 2) circle (\circlesize) node (C1) {};
        
        \draw[Colossus] (2.6, 2) node[rotate=45, right]{\color{black}{\;Colossus}};
        \filldraw[Colossus] (3.1, 2) circle (\circlesize) node (S1) {};
        
        \draw[black] (2.2, 2) node[rotate=45, right]{\color{black}{\;\;Great Lighthouse}};
        \filldraw[GreatLighthouse] (2.5, 2) circle (\circlesize) node (L1) {};
        
        \filldraw[Pyramids] (0.1, 2) circle (\circlesize) node[rotate=45, right] (Y1) {\color{black}{\;Pyramids}};
        
        \filldraw[TempleOfArtemis] (9.3, 1) circle (\circlesize) node (T2) {};
        \filldraw[Piraeus] (8.7, 1) circle (\circlesize) node (P2) {};
        \filldraw[HangingGardens] (7.6, 1) circle (\circlesize) node (H2) {};
        \filldraw[AppianWay] (6.7, 1) circle (\circlesize) node (A2) {};
        \filldraw[Sphinx] (6.0, 1) circle (\circlesize) node (S2) {};
        \filldraw[StatueOfZeus] (3.4, 1) circle (\circlesize) node (Z2) {};
        \filldraw[GreatLibrary] (4.3, 1) circle (\circlesize) node (B2) {};
        \filldraw[Mausoleum] (3.6, 1) circle (\circlesize) node (M2) {};
        \filldraw[CircusMaximus] (3.3, 1) circle (\circlesize) node (C2) {};
        \filldraw[Colossus] (1.8, 1) circle (\circlesize) node (S2) {};
        \filldraw[GreatLighthouse] (4.7, 1) circle (\circlesize) node (L2) {};
        \filldraw[Pyramids] (0.7, 1) circle (\circlesize) node (Y2) {};
        
        \draw[TempleOfArtemis, thick] (9.7, 2) to[out=-90,in=90] (9.3, 1);
        \draw[Piraeus, thick] (8.9, 2) to[out=-90,in=90] (8.7, 1);
        \draw[HangingGardens, thick] (7.6, 2) to[out=-90,in=90] (7.6, 1);
        \draw[AppianWay, thick] (6.7, 2) to[out=-90,in=90] (6.7, 1);
        \draw[Sphinx, thick] (6.0, 2) to[out=-90,in=90] (6.0, 1);
        \draw[StatueOfZeus, thick] (5.2, 2) to[out=-105,in=75] (3.4, 1);
        \draw[GreatLibrary, thick] (4.0, 2) to[out=-90,in=90] (4.3, 1);
        \draw[Mausoleum, thick] (3.3, 2) to[out=-90,in=90] (3.6, 1);
        \draw[CircusMaximus, thick] (3.2, 2) to[out=-90,in=90] (3.3, 1);
        \draw[Colossus, thick] (3.1, 2) to[out=-100,in=80] (1.8, 1);
        \draw[GreatLighthouse, thick] (2.5, 2) to[out=-75,in=105] (4.7, 1);
        \draw[Pyramids, thick] (0.1, 2) to[out=-90,in=90] (0.7, 1);
    \end{tikzpicture}
    \caption{Wonder preference of \AIname (top) and human players (bottom).}
    \label{fig:wonder-preference-comparison}
\end{figure}

\textbf{Wonder selection:}
Fig.~\ref{fig:wonder-preference-comparison} compares \AIname's wonder preference with that of human players on BGA \cite{bga}.
The 5 extra-turn wonders rank as \AIname's favorite, and their preference rates are nearly identical to humans.
On the other hand, the Pyramids are almost always selected last.
Among the remaining 6 wonders, \AIname favors the Statue of Zeus and dislikes the Great Lighthouse.
While human experts recognize the strength of the Statue of Zeus, such a strong preference seems in disagreement with how humans currently play 7WD.

\textbf{Building wonders in age I:}
\AIname builds an extra-turn wonder 52\% of the times when there are 2 cards left in age I, and 12\% of the times when there are 4 cards left.
This is a known tactic that allows a player to obtain more cards in age I and also start age II.
\AIname frequently employs this tactic even with a single extra-turn wonder left, while human players tend to be more conservative.

\AIname also appears to build wonders earlier than human players usually would.
For instance, in age I, \AIname builds an average of 0.83 wonders as \nth{1} player and 0.49 as \nth{2} player.

\textbf{Cards:}
Fig.~\ref{fig:top_cards} shows the cards most frequently built by \AIname in each age (which are not necessarily the cards that are built earlier).
In age I, the Tavern is the most built card, with production cards (brown and gray) being built more often than the remaining yellow cards.
Among resources, papyrus is preferred to glass, and wood is preferred to stone and clay.
\AIname often builds blue cards, even in age I and age II, to try to establish an early lead in victory points.

\begin{figure}[t]
    \centerline{\includegraphics[height=2.5cm]{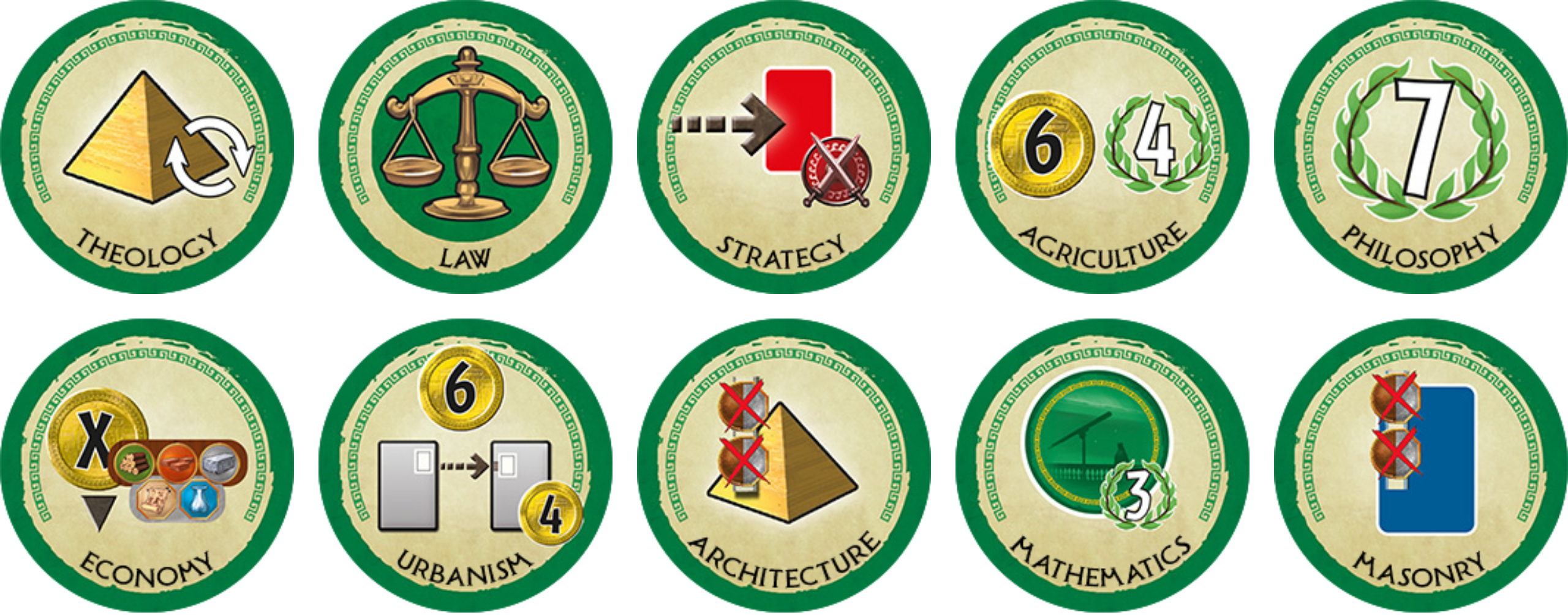}}
    \caption{Progress tokens ranked by frequency of selection in \AIname games. %
    }
    \label{fig:tokens}
\end{figure}

\textbf{Progress tokens:}
Fig.~\ref{fig:tokens} shows all progress tokens sorted by how frequently \AIname selects them.
As expected, the top 3 positions are held by Theology, Law, and Strategy, which are considered to be the most powerful tokens in the game.

\subsection{Balancing 7 Wonders Duel}

It is known in the community that the \nth{1} player has a significant advantage in 7WD. Self-play games by \AIname confirm this belief: out of 10k games, the \nth{1} player had a 66.8\% win rate (see also Fig.~\ref{fig:victory-type-plot}).
This is substantially more than the estimate of 55.7\% based on 605 games between human players on BGA \cite{bga}, further supporting the idea that \AIname is stronger than humans.
In the rest of this section, we propose rule variants that bring the win rate closer to $50\%$.

\textbf{Coin variant:}
A natural way to balance the game is to change the amount of coins given to each player during the setup.
We tried several options, estimating the win rate by generating 2k self-play games with the new setup, and computing the fraction of games won by the \nth{1} player (see Table \ref{table:coin-variant}, where the 95\% confidence intervals have a margin of error of $\sim$2\% for setups  other than 7-7).

Even a coin difference as large as 5 in favor of the \nth{2} player is not enough compensation. %
Indeed, a 1-coin change roughly corresponds to a 2.7\% change in the win rate. %

\begin{table}[tb]
\caption{Estimated win rate of the \nth{1} player in the coin variant}
    \centering
    {\small\renewcommand{\arraystretch}{1.4}%
    \begin{tabular}{cc|cccc}%
        & & \multicolumn{4}{c}{\textbf{\nth{2} player coins}} \\
        & & \textbf{7} & \textbf{8} & \textbf{9} & \textbf{10} \\
        \hline
        \multirow{3}{*}{\textbf{\makecell{\nth{1} player\\[.05cm]coins}}} & \textbf{7} & $66.8\%$ & $65.5\%$ & $63.4\%$ & $58.8\%$ \\
        & \textbf{6} & $63.7\%$ & $62.4\%$ & $59.2\%$ & $53.6\%$ \\
        & \textbf{5} & $62.5\%$ & $58.3\%$ & $59.5\%$ & $54.4\%$
    \end{tabular}%
    \vskip-0.45cm%
    }%
    \label{table:coin-variant}
\end{table}

\textbf{Wonder selection variants:}
The other variants we propose consist of altering the wonder selection phase, so that the advantage of starting age I is balanced by weaker wonders.

To estimate the win rate in each variant, we cannot directly generate self-play games since \AIname is not able to handle the new rules.
However, we can exploit the fact that afterstates reached at the end of the wonder selection phase are valid afterstates for the standard 7WD rules.
Specifically, we use \AIname's Transformer to estimate the value of all such possible afterstates;
then, for each variant, we use minimax to compute the optimal strategy during wonder selection (with the new rules), where the outcome is given by the value of the resulting afterstate.
With this method, we estimate a 67.6\% win rate for the \nth{1} player with the standard rules, which is fairly close to our aforementioned estimate of 66.8\%.

In these variants, the player making the first decision is not necessarily the same as the player starting age I.
Therefore, we refer to the two players as \emph{Aristotle} %
and \emph{Cleopatra}.
We report the estimated win rate for Cleopatra.%

\begin{itemize}
    \item \textbf{Variant 1} (60.0\% win rate): After seeing the first 4 wonders, Aristotle decides whether to start the first draft; the other player starts the second draft. Cleopatra starts age I.
    This variant is known within the 7WD community.
    
    \item \textbf{Variant 2} (54.6\% win rate): In both the first and second draft, after seeing the 4 wonders, Aristotle decides whether to start the draft. Cleopatra starts age I. The win rate is comparable to the expected score for white in Chess, deemed acceptable for competitive play.
    
    \item \textbf{Variant 3} (51.6\% win rate): Aristotle sees all 8 available wonders; he assigns 4 to the player starting age I and the remaining 4 to the other player. Cleopatra then decides whether to start age I. This can be considered an ``expert variant'', as optimal play requires an advanced understanding of the strengths of the wonders.
\end{itemize}

\section{Conclusion and Future Work}

In this paper, we have described our ongoing efforts in building, evaluating, and using \AIname to understand and improve 7WD.
Going forward, we intend to further assess \AIname's strengths and weaknesses, discover deeper insights into 7WD strategy, and playtest ideas to balance the game.

\section*{Acknowledgment}

The authors thank Francesco Bardi, Ugo Bindini, Michele Borassi, Federico Poli, Jayson Rickel, Joël Seytre, Enrico Toffoli, Marco Trevisiol, and Jean-Marc Trinh for the discussions and for playing against \AIname.

\bibliographystyle{IEEEtran}
\bibliography{bibliography}

\begin{thebibliography}{1}
\providecommand{\url}[1]{#1}
\csname url@samestyle\endcsname
\providecommand{\newblock}{\relax}
\providecommand{\bibinfo}[2]{#2}
\providecommand{\BIBentrySTDinterwordspacing}{\spaceskip=0pt\relax}
\providecommand{\BIBentryALTinterwordstretchfactor}{4}
\providecommand{\BIBentryALTinterwordspacing}{\spaceskip=\fontdimen2\font plus
\BIBentryALTinterwordstretchfactor\fontdimen3\font minus
  \fontdimen4\font\relax}
\providecommand{\BIBforeignlanguage}[2]{{%
\expandafter\ifx\csname l@#1\endcsname\relax
\typeout{** WARNING: IEEEtran.bst: No hyphenation pattern has been}%
\typeout{** loaded for the language `#1'. Using the pattern for}%
\typeout{** the default language instead.}%
\else
\language=\csname l@#1\endcsname
\fi
#2}}
\providecommand{\BIBdecl}{\relax}
\BIBdecl

\bibitem{silver2018general}
D.~Silver, T.~Hubert, J.~Schrittwieser, I.~Antonoglou, M.~Lai, A.~Guez,
  M.~Lanctot, L.~Sifre, D.~Kumaran, T.~Graepel, T.~Lillicrap, K.~Simonyan, and
  D.~Hassabis, ``A general reinforcement learning algorithm that masters chess,
  shogi, and {Go} through self-play,'' \emph{Science}, vol. 362, no. 6419, pp.
  1140--1144, 2018.

\bibitem{vaswani2017attention}
A.~Vaswani, N.~Shazeer, N.~Parmar, J.~Uszkoreit, L.~Jones, A.~N. Gomez,
  {\L}.~Kaiser, and I.~Polosukhin, ``Attention is all you need,''
  \emph{Advances in Neural Information Processing Systems}, vol.~30, 2017.

\bibitem{chatgpt}
OpenAI, ``Introducing {ChatGPT},'' \url{https://openai.com/blog/chatgpt}, 2022,
  accessed 2024-04-24.

\bibitem{bga}
Le007n, ``{7WD} in numbers,''
  \url{https://forum.boardgamearena.com/viewtopic.php?t=25369}, 2022, accessed
  2024-04-24.

\end{thebibliography}

\end{document}